\documentclass[conference]{IEEEtran}
\IEEEoverridecommandlockouts
\usepackage{amsmath,amssymb,amsfonts}
\usepackage{algorithm}
\usepackage{graphicx}
\usepackage{textcomp}
\usepackage{xcolor}
\usepackage{booktabs}
\usepackage{multirow}
\usepackage{subfigure}
\usepackage{pifont}
\usepackage{xspace}
\usepackage{hyperref}
\usepackage{algpseudocode}
\usepackage{color}
\usepackage{bm}
\usepackage{tabularx}
\usepackage{bm}
\usepackage[square,sort,comma,numbers]{natbib}
\usepackage{caption}

\hypersetup{
    colorlinks=true,
}
\def\BibTeX{{\rm B\kern-.05em{\sc i\kern-.025em b}\kern-.08em
    T\kern-.1667em\lower.7ex\hbox{E}\kern-.125emX}}
\begin{document}

\title{
    Masked Autoencoders are Parameter-Efficient Federated Continual Learners\\
    \thanks{This work was supported by the STCSM (22QB1402100) and NSFC (62231019, 12071145). Corresponding author: Xiangfeng Wang.}
}

\author{
    \IEEEauthorblockN{
    Yuchen HE\IEEEauthorrefmark{1},
    Xiangfeng WANG\IEEEauthorrefmark{2}\IEEEauthorrefmark{1}
    }
    \IEEEauthorblockA{
    \IEEEauthorrefmark{1}School of Computer Science and Technology, East China Normal University, Shanghai, China 200062\\
    \IEEEauthorrefmark{2}Shanghai Formal-Tech Information Technology Co., Lt, Shanghai, China 200062
    }
}

\maketitle

\begin{abstract}

    Federated learning is a specific distributed learning paradigm in which a central server aggregates updates from multiple clients' local models,
    thereby enabling the server to learn without requiring clients to upload their private data, maintaining data privacy.
    While existing federated learning methods are primarily designed for static data,
    real-world applications often require clients to learn new categories over time.
    This challenge necessitates the integration of continual learning techniques, leading to federated continual learning (FCL).
    To address both catastrophic forgetting and non-IID issues,
    we propose to use masked autoencoders (MAEs) as parameter-efficient federated continual learners, called \textbf{pMAE}.
    pMAE learns reconstructive prompt on the client side through image reconstruction using MAE.
    On the server side, it reconstructs the uploaded restore information to capture the data distribution across previous tasks and different clients,
    using these reconstructed images to fine-tune discriminative prompt and classifier parameters tailored for classification,
    thereby alleviating catastrophic forgetting and non-IID issues on a global scale.
    Experimental results demonstrate that pMAE achieves performance comparable to existing prompt-based methods
    and can enhance their effectiveness, particularly when using self-supervised pre-trained transformers as the backbone.
    Code is available at: \url{https://github.com/ycheoo/pMAE}.

\end{abstract}


\section{Introduction}
\label{sec:intro}

\begin{figure}[htp!]
    \centering
    \includegraphics[width=0.99\linewidth]{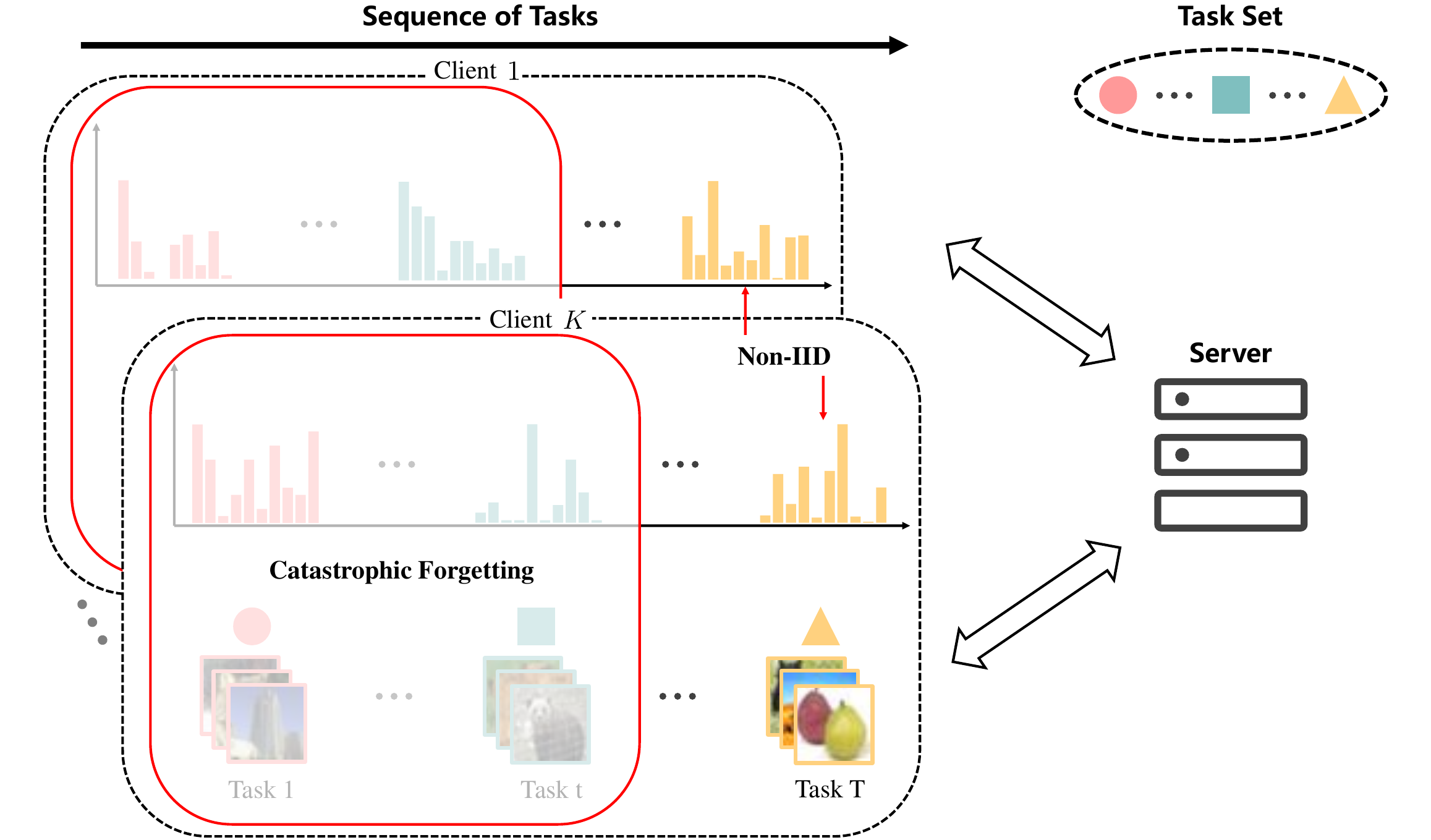}
    \caption{
        Illustration of federated continual learning (FCL), wherein clients sequentially learn over $T$ tasks. 
        Each client continuously updates its local model with class-imbalanced private data specific to each task, 
        and further transfers the well-chosen parameters to server for aggregation. 
        The aggregated global model needs to maintain discriminability across all observed classes within the task set.
    }
    \label{fig:FCL_setting}
\end{figure}

Federated learning~\cite{mcmahan2017communication,yang2019federated,kairouz2021advances,huang2024federated} is a privacy-preserving decentralized learning paradigm that enables collaborative model training without requiring clients to upload their private data.
Traditional federated learning assumes static data, but in real-world scenarios, clients' data is often dynamic, with old task data being removed and new task data emerging.
To address these dynamic scenarios, continual learning~\cite{de2022continual,masana2023class,wang2024comprehensive,zhou2024continual} techniques need to be integrated into federated learning,
a combination known as federated continual learning (FCL)~\cite{yoon2021federated,dong2022fcil}, as shown in Figure~\ref{fig:FCL_setting}.

In FCL scenarios, new task data is continuously generated on clients, while old task data cannot be fully utilized for model updates.
This limitation results in catastrophic forgetting~\cite{kirkpatrick2017overcoming,rebuffi2017icarl},
where the model's performance on previous tasks degrades significantly after training on new tasks.
Additionally, due to the non-IID (non-independent and identically distributed)~\cite{zhao2018federated,li2022federated} nature of FCL,
local data distributions across clients may vary significantly, resulting in a suboptimal global model~\cite{li2022federated,zhang2022federated}.

The performance drop caused by catastrophic forgetting is due to the absence of old task data,
which leads the model to overfit the data of current task~\cite{kirkpatrick2017overcoming,rebuffi2017icarl}.
Similarly, the performance drop caused by non-IID arises because the local model fits the non-IID data of individual clients,
preventing the aggregation of a global model that reflects the overall data distribution~\cite{li2022federated,zhang2022federated}.
If we could effectively capture the distribution of old task data and account for the differing data distributions across clients,
both catastrophic forgetting and the non-IID problem could be significantly alleviated.

Building on the insights from~\citet{zhai2023masked}, we propose to use masked autoencoders (MAEs) as parameter-efficient federated continual learners, called pMAE.
Originally introduced as scalable vision learners, MAEs~\cite{he2022mae} leverage self-supervised learning~\cite{chen2021mocov3,he2022mae,zhou2022ibot} to learn effective visual representations.
Instead of focusing solely on representation learning, pMAE emphasizes the ability of MAEs to reconstruct images from restore information.
In pMAE, client-side learning not only prioritizes classification as the primary task but also focuses on reconstruction as a secondary task.
On the server side, the images reconstructed from uploaded restore information are utilized to fine-tune model parameters tailored for classification.
Reconstructed images represent distributions from old tasks and diverse client data, thereby enabling the global mitigation of catastrophic forgetting and non-IID issues.

In FCL, the limitations of client-side computational resources make it burdensome to train a transformer-based model (such as MAEs)~\cite{vaswani2017attention,devlin2019bert,dosovitskiy2021vit} from scratch;
therefore, we employ pre-trained transformers~\cite{han2021pre} as the backbone.
To further enhance parameter efficiency, we freeze the parameters of the pre-trained transformer and leverage prompt tuning~\cite{lester2021power,jia2022vpt}.
Specifically, clients utilize discriminative prompt and classifier parameters for classification while employing reconstructive prompt for reconstructing masked images.
After completing training, clients randomly extract restore information from a subset of the images and upload them to the server.
The server then reconstructs images using reconstructive prompt and restore information,
fine-tuning discriminative prompt and classifier parameters in the process, which mitigates performance degradation from catastrophic forgetting and non-IID issues on a global scale.
Given the generality of our approach, it can be directly integrated with existing prompt-based methods~\cite{wang2022l2p, wang2022dualprompt, smith2023coda-prompt} to enhance their performance.

Our contributions can be summarized as follows:

\noindent - We propose to use masked autoencoders (MAEs) for FCL tasks, mitigating catastrophic forgetting and non-IID issues globally by reconstructing images on the server.

\noindent - We introduce prompt tuning as a parameter-efficient method for fine-tuning pre-trained transformers, employing discriminative prompt for the classification task and reconstructive prompt for the reconstruction task.

\noindent - Experimental results indicate that pMAE achieves performance comparable to existing prompt-based methods and can be directly integrated to enhance their effectiveness, particularly when utilizing self-supervised pre-trained transformers as the backbone.

\section{Related Work}
\label{sec:rw}

\subsection{Continual Learning}

Continual learning, also known as incremental learning or lifelong learning~\cite{de2022continual,masana2023class,wang2024comprehensive,zhou2024continual},
is a learning paradigm in which models acquire knowledge from a series of sequential tasks.
This approach is typically categorized into three main types:
Regularization-based methods~\cite{kirkpatrick2017overcoming,li2018lwf,smith2023closer}, which limit changes to model parameters during the learning of new tasks;
Rehearsal-based methods~\cite{rebuffi2017icarl,wu2019bic,zhai2023masked}, which involve revisiting historical data from previous tasks while learning new ones;
Model expansion-based methods~\cite{yoon2018lifelong,yan2021der,zhou2023memo}, which incorporate additional network structures when acquiring new tasks.

Recent advances in continual learning have introduced approaches that leverage knowledge from pre-trained transformers~\cite{vaswani2017attention,devlin2019bert,dosovitskiy2021vit}
through parameter-efficient tuning~\cite{lester2021power,li2021prefix,jia2022vpt,hu2022lora,chen2022adaptformer}.
In particular, prompt-based methods~\cite{wang2022l2p,wang2022dualprompt,smith2023coda-prompt,wang2023hide,li2024steering} instruct pre-trained transformers using lightweight prompts~\cite{lester2021power,li2021prefix,jia2022vpt}.
EASE~\cite{zhou2024expandable} trains a unique adapter module~\cite{chen2022adaptformer} for each task, aiming to develop task-specific subspaces.
ADAM~\cite{zhou2024revisiting} and LAE~\cite{gao2023lae} propose general continual learning frameworks, which can be smoothly integrated with other parameter-efficient tuning methods.

\subsection{Federated Continual Learning}

Federated learning is a decentralized learning paradigm where multiple clients jointly build a global model without exchanging their personal data~\cite{mcmahan2017communication,yang2019federated,kairouz2021advances,huang2024federated}.
Nevertheless, this learning paradigm typically presupposes that clients' data does not change, overlooking the reality that data tends to evolve and grow over time.

Federated continual learning (FCL)~\cite{yoon2021federated,dong2022fcil} seeks to tackle the challenge of enabling clients to learn progressively across multiple tasks by integrating continual learning techniques into the federated learning framework.
FCL faces obstacles such as catastrophic forgetting\cite{kirkpatrick2017overcoming,rebuffi2017icarl}, a known issue in continual learning, and the non-IID issue~\cite{zhao2018federated,li2022federated} common in federated learning.

In recent years, the field of FCL has experienced significant growth~\cite{yoon2021federated,dong2022fcil,qi2023better,zhang2023target}.
With the rise of pre-trained models~\cite{han2021pre}, FCL approaches leveraging parameter-efficient tuning~\cite{lester2021power,li2021prefix,jia2022vpt,hu2022lora,chen2022adaptformer}
on pre-trained transformers~\cite{vaswani2017attention,devlin2019bert,dosovitskiy2021vit} have demonstrated impressive performance while reducing communication overhead.
Fed-CPrompt~\cite{bagwe2023fedcprompt}, employing prompting techniques~\cite{lester2021power,li2021prefix,jia2022vpt}, enables communication-efficient FCL without the need for data rehearsal.
Powder~\cite{piao2024federated} is designed to promote effective knowledge transfer encapsulated in prompts across sequentially learned tasks and different clients.
PILoRA~\cite{guo2024pilora}, which incorporates LoRA~\cite{hu2022lora} and prototype~\cite{snell2017proto}, enhances representation learning and utilizes heuristic information between prototypes and class features.

\section{Preliminaries}
\label{sec:pre}

\subsection{Problem Definition}

In FCL~\cite{yoon2021federated,dong2022fcil}, a server oversees a group of $K$ clients that sequentially learn across $T$ tasks.
Training data is randomly assigned to clients and cannot be shared or uploaded.
This work primarily focuses on addressing the challenges of class-incremental learning~\cite{rebuffi2017icarl} within the context of continual learning.
Class-incremental learning involves handling $T$ sets of non-overlapping, class-labeled datasets,
denoted as $\{ \mathcal{D}^1, \cdots, \mathcal{D}^T\ :\ \mathcal{D}^t = {(x_i^t, y_i^t)}_{i=1}^{n^t} \}$,
where $\mathcal{D}^t$ is the training dataset for the $t$-th task, and $y_i^t \in \mathcal{Y}^t$ refers to the class label corresponding to $x_i^t$.
For any two distinct tasks, the class sets are mutually exclusive, represented as $\mathcal{Y}^{t} \cap \mathcal{Y}^{t'} = \emptyset$ for $t \neq t'$.
During the training of task $t$, updates can only be made using $\mathcal{D}^t$ or a limited amount of rehearsal data~\cite{rebuffi2017icarl}.
The updated model must classify all classes encountered across the first $t$ tasks, i.e., $\cup_{i=1}^{t} \mathcal{Y}^i$.

When extending class-incremental learning to a federated learning setting,
the training data for the $t$-th task is distributed randomly across $K$ clients,
represented as $\mathcal{D}^t = \{{\mathcal{D}_{k}^{t}} \}_{k=1}^{K}$.
For any two clients $k \neq k'$, their datasets are disjoint,
meaning $\mathcal{D}_{k}^t \cap \mathcal{D}_{k'}^t = \emptyset$.

\subsection{Masked Autoencoders}

Masked Autoencoders (MAEs) were originally introduced as scalable vision learners in a self-supervised paradigm by reconstructing images from masked inputs~\cite{he2022mae}.

\begin{figure*}[htp!]
    \centering
    \includegraphics[width=0.95\textwidth]{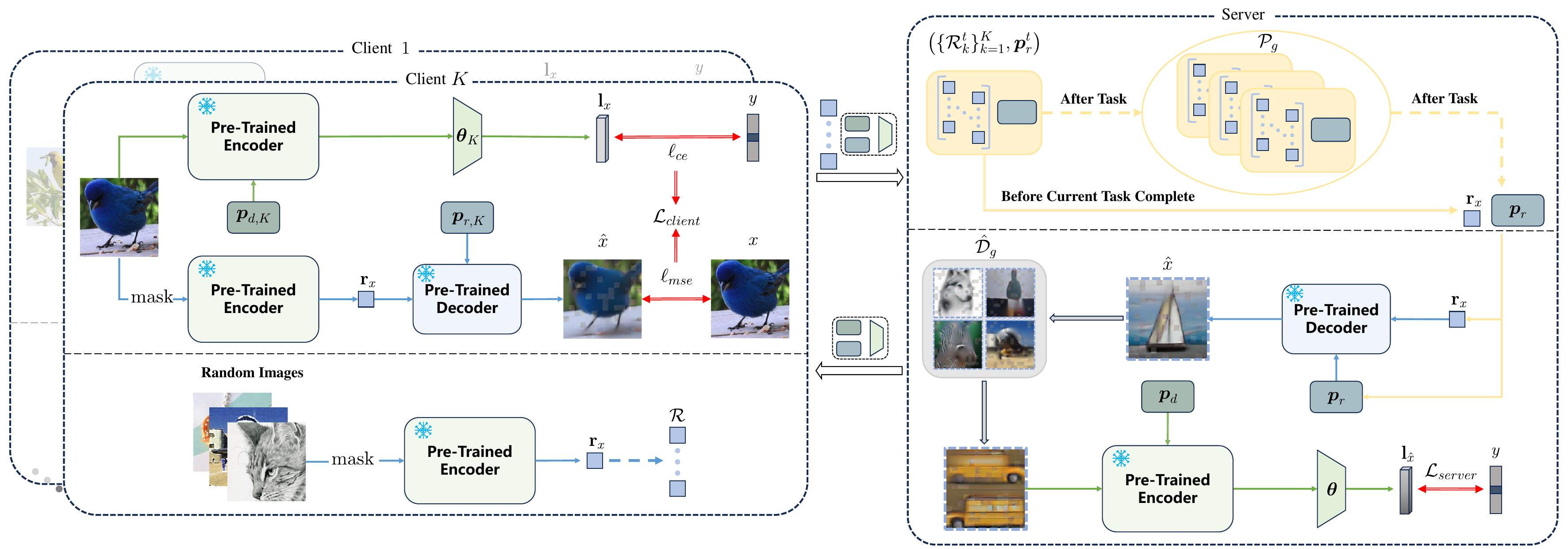}
    \caption{
        The overview of proposed pMAE framework.
        1) Client: Extract labeled restore information using Eq.~\eqref{eq:encoder}
        and optimize prompts and classifier parameters of the local model by $\mathcal{L}_{client}$.
        2) Server: Generate reconstructed images using Eq.~\eqref{eq:decoder}
        and fine-tune discriminative prompt and classifier parameters of the aggregated global model by $\mathcal{L}_{server}$.
        3) Lightweight prompts, classifier parameters and labeled restore information are transmitted between clients and the server,
        while no additional data is stored on each client.
        4) Images are reconstructed into tensors and subsequently used for fine-tuning, ensuring that \textbf{no real images are stored on the server}.
    }
    \label{fig:overview}
\end{figure*}

The MAE encoder is a Vision Transformer (ViT) model~\cite{dosovitskiy2021vit},
where input images are divided into non-overlapping patches,
and a significant portion (e.g., 75 \%) of these patches are randomly masked.
The remaining unmasked patches are encoded into visible tokens and passed to the decoder along with restore ids,
which indicate the positions of masked tokens.

The MAE decoder, which is used only for reconstruction task during the pre-training stage, is designed much narrower and shallower than the encoder.
The visible tokens from the encoder are combined with learnable mask tokens that replace the masked patches based on restore ids.
The decoder then processes the combination of visible tokens and mask tokens to reconstruct the original input images in the pixel space.

\subsection{Prompt Tuning}
\label{subsection:pt}

Prompt tuning~\cite{lester2021power,jia2022vpt} is a parameter-efficient method for tuning pre-trained transformers~\cite{vaswani2017attention,devlin2019bert,dosovitskiy2021vit},
where a prompt parameter $\boldsymbol{p} \in \mathbb{R}^{L_{\boldsymbol{p}} \times D}$,
with $L_{\boldsymbol{p}}$ representing the sequence length and $D$ denoting the embedding dimension,
is prepended to the encoded feature $\boldsymbol{h}$ of input $x$.
Specifically, an identical $\boldsymbol{p}$ is prepended to the query, key, and value of encoded feature $\boldsymbol{h}$,
denoted as $\boldsymbol{h}_Q$, $\boldsymbol{h}_K$, and $\boldsymbol{h}_V$, respectively.
This prepending occurs within the multi-head self-attention (MSA) layers and is expressed as:
\begin{equation}\label{eq:prompt_tuning}
    {\rm{MSA}}([\boldsymbol{p};\boldsymbol{h}_Q], [\boldsymbol{p};\boldsymbol{h}_K], [\boldsymbol{p};\boldsymbol{h}_V])
\end{equation}
where $[\cdot \, ; \cdot]$ denotes the concatenation operation along the sequence length.

In pMAE, we introduce a discriminative prompt for classification and a reconstructive prompt for image reconstruction.
Following prior works~\cite{wang2022l2p,wang2022dualprompt,smith2023coda-prompt},
the discriminative prompt length is set to $L_{\boldsymbol{p}}=20$ and is inserted into the first five transformer blocks with individual parameters.
The reconstructive prompt is set with $L_{\boldsymbol{p}}=5$ and is only inserted into the first layer, which we designed specifically.

\section{Methodology of the pMAE}
\label{sec:method}

In FCL, the performance is influenced by two key issues: catastrophic forgetting and non-IID,
which are caused by the absence of old task data and varying local data distributions.
To tackle catastrophic forgetting and non-IID issues in a parameter-efficient manner,
pMAE utilizes the reconstruction ability of MAEs and the instruction ability of prompt tuning.

On the client side,
pre-trained transformers learn from prompt tuning,
where discriminative prompt enhance the classification performance of the encoder,
and reconstructive prompt improve the reconstruction performance of the decoder.
Restore information, along with tuned model parameters, are then extracted and uploaded to the server.

On the server side,
the received restore information and reconstructive prompt are used to reconstruct images,
which are subsequently used to fine-tune the discriminative prompt and classifier parameters,
mitigating catastrophic forgetting and non-IID issues from a global perspective.

\subsection{Client Side: Learning from Prompting}

The backbone of the local model consists of a pre-trained transformer encoder $F_{\Phi}$ and a pre-trained transformer decoder $G_{\psi}$, with the model parameters $\Phi$ and $\psi$ remaining frozen.
To instruct pre-trained transformers and obtain output logits for classification, tunable prompts and classifier are introduced.

Specifically, for the classification task, there are a discriminative prompt $\boldsymbol{p}_{d}$ and a classifier $H_{\boldsymbol{\theta}}$ parameterized by $\boldsymbol{\theta}$.
As described in Sec~\ref{subsection:pt}, prompt $\boldsymbol{p}$ is prepended to the encoded feature $\boldsymbol{h}$ of input $x$.
For convenience, we use $[\cdot \, , \cdot]$ to denote such combination.
The output logits can then be denoted as:
\begin{equation}\label{eq:ox}
    \mathbf{l}_{x} = H_{\boldsymbol{\theta}}(F_{\Phi} \left( \left[ x , \boldsymbol{p}_{d} \right] \right)),
\end{equation}
and the classification loss is computed using the cross-entropy (CE) between the ground truth $y$ and the output logits $\mathbf{l}_{x}$, denoted as $\ell_{ce}(\mathbf{l}_{x}, y)$.

Similarly, for the reconstruction task, we have a reconstructive prompt $\boldsymbol{p}_{r}$,
which is utilized to enhance the decoder's reconstruction performance.
During the learning process, input images $x$ are masked with a portion of 75 \% and encoded by the pure encoder $F_{\Phi}$ without discriminative prompt $\boldsymbol{p}_{d}$ as:
\begin{equation}\label{eq:encoder}
    \mathbf{r}_{x} = F_{\Phi} \left( \mathrm{mask}(x) \right),
\end{equation}
where $\mathbf{r}_{x}$ represents the restore information, which includes the encoded visible tokens and the restore ids of the masked input images $\mathrm{mask}(x)$.

The decoder, $G_{\psi}$, subsequently outputs the reconstructed images with the reconstructive prompt $\boldsymbol{p}_{r}$ as:
\begin{equation}\label{eq:decoder}
    \hat{x} = G_{\psi} \left( \left[ \mathbf{r}_{x} , \boldsymbol{p}_{r} \right] \right),
\end{equation}
where $\hat{x}$ represents the reconstructed version of the original images $x$,
and the loss of reconstruction is computed as the mean squared error (MSE) between the reconstructed and original images in pixel space,
denoted by $\ell_{mse}(\hat{x}, x)$.

To summarize the modeling process above, the overall loss function for each client is:
\begin{equation}\label{eq:l_client}
    {\textstyle{\mathcal{L}_{client}(\boldsymbol{p}_{d}, \boldsymbol{\theta}, \boldsymbol{p}_{r}) = \sum_{(x,y) \in \mathcal{D}_{k}^{t}} \left( \ell_{ce}(\mathbf{l}_{x}, y) + \ell_{mse}(\hat{x}, x) \right).}}
\end{equation}
where $\mathcal{D}_{k}^{t}$ represents the private dataset of client $k$ at task $t$.

Once the training process is complete, clients will randomly selects a number of $u$ images and extracts their labeled restore information, represented as $\mathcal{R} = \{ (\mathbf{r}_{x_i}, y_i) \}_{i=1}^{u}$, which will be uploaded to the server for global image reconstruction.
Ultimately, clients uploads only the prompts $\boldsymbol{p}_{d}$ and $\boldsymbol{p}_{r}$, the classifier parameters $\boldsymbol{\theta}$, and the labeled restore information $\mathcal{R}$ to the server, which are significantly more lightweight than the model parameters of encoder and decoder.
For clarity of notation, transmitted model parameters are collectly expressed as $\boldsymbol{w} = \left\{ \boldsymbol{p}_{d}, \boldsymbol{\theta}, \boldsymbol{p}_{r} \right\}$.

\subsection{Server Side: Fine-tuning from Reconstructing}

The discriminative prompt $\boldsymbol{p}_{d}$ and classifier parameters $\boldsymbol{\theta}$ uploaded by the client can be aggregated by FedAvg~\cite{mcmahan2017communication} to form a global model for classification.
However, the performance of the global model may be compromised due to catastrophic forgetting caused by the absence of old task data, as well as non-IID issue arising from varying local distributions.
Therefore, it is essential to utilize the reconstructive prompt $\boldsymbol{p}_{r}$ and the labeled restore information $\mathcal{R}$ to capture distributions from old tasks and diverse clients.

For the labeled restore information uploaded by clients, the server will aggregates it into $\left\{ \mathcal{R}_{k} \right\}_{k=1}^{K}$,
which, along with the FedAvg-aggregated reconstructive prompt $\boldsymbol{p}_{r}$, is denoted as $\left( \{{\mathcal{R}}_k\}_{k=1}^{K}, \boldsymbol{p}_{r} \right)$ for image reconstruction.
Since $\left\{ \mathcal{R}_{k} \right\}_{k=1}^{K}$ contains distribution information from different clients, it can effectively mitigate the non-IID issue.
Specifically, based on Eq.~\eqref{eq:decoder}, the reconstructive prompt $\boldsymbol{p}_{r}$ is used to reconstruct the labeled image $(\hat{x}, y)$ with labeled restore information $(\mathbf{r}_{x}, y)$ from $\left\{ \mathcal{R}_{k} \right\}_{k=1}^{K}$,
creating a reconstructed global dataset $\hat{\mathcal{D}}_{g}$.

\begin{algorithm}[htp!] 
	\renewcommand{\algorithmicrequire}{\textbf{Input:}}
	\renewcommand{\algorithmicensure}{\textbf{Output:}}
	\caption{The pMAE Framework} 
	\label{alg1} 
    {\bf Input:}
    communication rounds per task $R$, local epochs $E$, number of clients $K$, private datasets $\{ \mathcal{D}_k^t \}_{k=1}^{K}$.
	\begin{algorithmic}[1]
    \Procedure{pMAE}{}
        \State 
        Initialize $\boldsymbol{w} = \{ \boldsymbol{p}_{d}, \boldsymbol{\theta}, \boldsymbol{p}_{r} \}$,
        restore pool $\mathcal{P}_{g} = \{\}$;
        \For{each task $t = 1, 2, \cdots$}
            \For{$\tau = 1, 2, \cdots, R$}
                \For{$k = 1, 2, \cdots, K$ {\bf in parallel}}
                    \State $\boldsymbol{w}_{k}^{\tau+1}, \mathcal{R}_{k}^t \leftarrow $ \textproc{LocalUpdate} $(t, k, {\boldsymbol{w}}^{r})$;
                \EndFor
                \State $\boldsymbol{w}^{\tau+1} \leftarrow \sum_{k=1}^K \frac{| \mathcal{D}_k^t |}{| \mathcal{D}^t |} \boldsymbol{w}_{k}^{\tau+1}$;
                \If{$\tau \neq R$}
                    \State Generate $\hat{\mathcal{D}}_{g}$ with $\left( \{{\mathcal{R}}_k^t\}_{k=1}^{K}, \boldsymbol{p}_{r}^{\tau+1} \right)$;
                    \State Optimize $\boldsymbol{p}_{d}^{\tau+1}, \boldsymbol{\theta}^{\tau+1}$ by Eq.~\eqref{eq:l_server};
                \EndIf
            \EndFor
            \State $\mathcal{P}_{g} \leftarrow \mathcal{P}_{g} \cup \left( \{{\mathcal{R}}_k^t\}_{k=1}^{K}, \boldsymbol{p}_{r}^{t} \right)$;
            \State Generate $\hat{\mathcal{D}}_{g}$ with $\mathcal{P}_{g} = \left\{ \left( \{{\mathcal{R}}_k^i\}_{k=1}^{K}, \boldsymbol{p}_{r}^{i} \right) \right\}_{i=1}^{t}$;
            \State Optimize $\boldsymbol{p}_{d}^{t}, \boldsymbol{\theta}^{t}$ by Eq.~\eqref{eq:l_server};
        \EndFor
    \EndProcedure
    \Statex
    \Function{LocalUpdate}{$t, k, {\boldsymbol{w}}$}
        \State $\boldsymbol{w}_k = \{ \boldsymbol{p}_{d, k}, \boldsymbol{\theta}_{k}, \boldsymbol{p}_{r, k} \} \leftarrow \boldsymbol{w}$;
        \For{$e = 1, 2, \cdots, E$}
            \For{$(x, y)$ in $\mathcal{D}_k^t$}
                \State
                Optimize $\boldsymbol{w}_k$ by Eq.~\eqref{eq:l_client};
            \EndFor
        \EndFor
        \State
        Extract labeled restore information $\mathcal{R}_k^t$ by Eq.~\eqref{eq:encoder};
        \State \textbf{return} $\boldsymbol{w}_k, \mathcal{R}_k^t$.
    \EndFunction
    \end{algorithmic} 
\end{algorithm}

This reconstructed global dataset is then used to fine-tune the discriminative prompt and classifier parameters as:
\begin{equation}\label{eq:l_server}
    {\textstyle{\mathcal{L}_{server}(\boldsymbol{p}_{d}, \boldsymbol{\theta}) = \sum_{(\hat{x},y) \in \hat{\mathcal{D}}_{g}} \ell_{ce}(\mathbf{l}_{\hat{x}}, y).}}
\end{equation}

However, due to the absence of old task data, the issue of catastrophic forgetting remains unresolved.
Therefore, we propose using a restore pool $\mathcal{P}_{g}$ to save past data distributions.
Specifically, after task $t$ is completed, the aggregated labeled restore information and reconstructive prompt are merged into the restore pool as:
$\mathcal{P}_{g} \leftarrow \mathcal{P}_{g} \cup \left( \{{\mathcal{R}}_k^t\}_{k=1}^{K}, \boldsymbol{p}_{r}^{t} \right)$.
Here, $\{{\mathcal{R}}_k^t\}_{k=1}^{K}$ and $\boldsymbol{p}_{r}^{t}$ are the aggregated labeled restore information and reconstructive prompt after the completion of task $t$,
which encompass the data distribution of task $t$.

During the subsequent server fine-tuning, if the current task is still ongoing,
$\left( \{{\mathcal{R}}_k\}_{k=1}^{K}, \boldsymbol{p}_{r} \right)$ is utilized for image reconstruction.
If the current task has ended, the aggregated labeled restore information and reconstructive prompt are merged into the restore pool $\mathcal{P}_{g}$.
We then use $\mathcal{P}_{g} = \left\{ \left( \{{\mathcal{R}}_k^i\}_{k=1}^{K}, \boldsymbol{p}_{r}^{i} \right) \right\}_{i=1}^{t}$,
which contains both old task data and current task data, to generate a reconstructed global dataset $\hat{\mathcal{D}}_{g}$ and use Eq.~\eqref{eq:l_server} to fine-tune the discriminative prompt and classifier parameters.

\section{Experiments}
\label{sec:exp}

\subsection{Experimental Setup}

\noindent\textbf{Datasets and Non-IID Setting.}
To evaluate the performance of pMAE in FCL scenarios, two representative benchmark datasets are utilized: CUB-200~\cite{wah2011caltech} and ImageNet-R~\cite{hendrycks2021many}.
These datasets are randomly divided into 20 tasks ($T=20$), with each task containing an equal number of distinct classes.
The CUB-200 dataset comprises approximately 12,000 images spanning 200 fine-grained bird categories.
The ImageNet-R dataset includes 200 classes with 24,000 training images and 6,000 testing images,
with all considered out-of-distribution data, despite sharing some classes with ImageNet~\cite{deng2009imagenet}, the dataset used for pre-training.
Consequently, ImageNet-R is widely used in recent continual learning research based on pre-trained models~\cite{wang2022dualprompt,smith2023coda-prompt}.

To simulate a non-IID environment, the Dirichlet distribution is applied,
a standard approach in federated learning~\cite{li2022federated,zhang2022federated}.
The degree of non-IIDness is controlled by the parameter $\beta$,
with lower values indicating a greater degree of non-IID distribution.
Notably, by employing this partitioning strategy, the private data for each client may include majority, minority, or missing classes, which could reflect real-world complexities.

\noindent\textbf{Evaluation Metric.}
In evaluating method performance, the metrics of accuracy $A_t$ and average accuracy $\bar{A}$ are employed,
as established in~\cite{chaudhry2018riemannian} and widely adopted in continual learning~\cite{wang2022l2p,wang2022dualprompt,smith2023coda-prompt}.
The accuracy on task $i$ after learning task $t$ is denoted as $a_{i, t}$,
and $A_t = \frac{1}{t} \sum_{i=1}^{t} a_{i, t}$ represents the accuracy of all tasks after learning task $t$.
The overall performance after task $T$ is quantified through the calculation of average accuracy$\bar{A} = \frac{1}{T} \sum_{t=1}^{T} A_{t}$,
where $T$ signifies the total number of tasks, thus providing a comprehensive measure across the entire learning progression.

\noindent\textbf{Baselines.}
Our pMAE is compared with three continual learning methods that utilize prompts:
L2P~\cite{wang2022l2p}, DualPrompt~\cite{wang2022dualprompt}, and CODA-Prompt~\cite{smith2023coda-prompt}.
Notably, these continual learning techniques will be adapted to the FCL setting,
referred to as FedL2P, FedDualP, and FedCODA-P, respectively.

\noindent\textbf{Implementation Details.}\phantomsection\label{impdetails}
The Adam optimizer~\cite{kingma2014adam} is applied across all methods,
utilizing a consistent batch size of 128 and a learning rate of $lr = 1e^{-3}$.
The number of clients is set to $K=10$, with a total of $R_{all}=200$ communication rounds and local update epochs $E=5$.
For each task, communication rounds are determined as $R = {\textstyle{\frac{R_{all}}{T}}}$.
For pMAE, the uploaded number of restore information is set to $u=4$, and fine-tuning on the server is performed over $E_{server}=5$ epochs.

All experiments are conducted across three random seeds 2023, 2024, and 2025, with the final results derived from them.

\noindent\textbf{Pre-training of Decoder.}\phantomsection\label{pre-training}
Previous works~\cite{wang2022l2p,wang2022dualprompt,smith2023coda-prompt} all use a ViT-B/16 model pre-trained on ImageNet~\cite{deng2009imagenet} as a supervised learning (Sup) backbone.
We also consider iBOT~\cite{zhou2022ibot}, which employs a self-supervised learning paradigm~\cite{chen2021mocov3,he2022mae,zhou2022ibot}.
Since our pMAE requires both an encoder for representation and a decoder for reconstruction, it necessitates pre-training a suitable decoder on both Sup and iBOT pre-trained encoders.

\begin{figure}[htp!]
    \centering
    \includegraphics[width=0.95\linewidth]{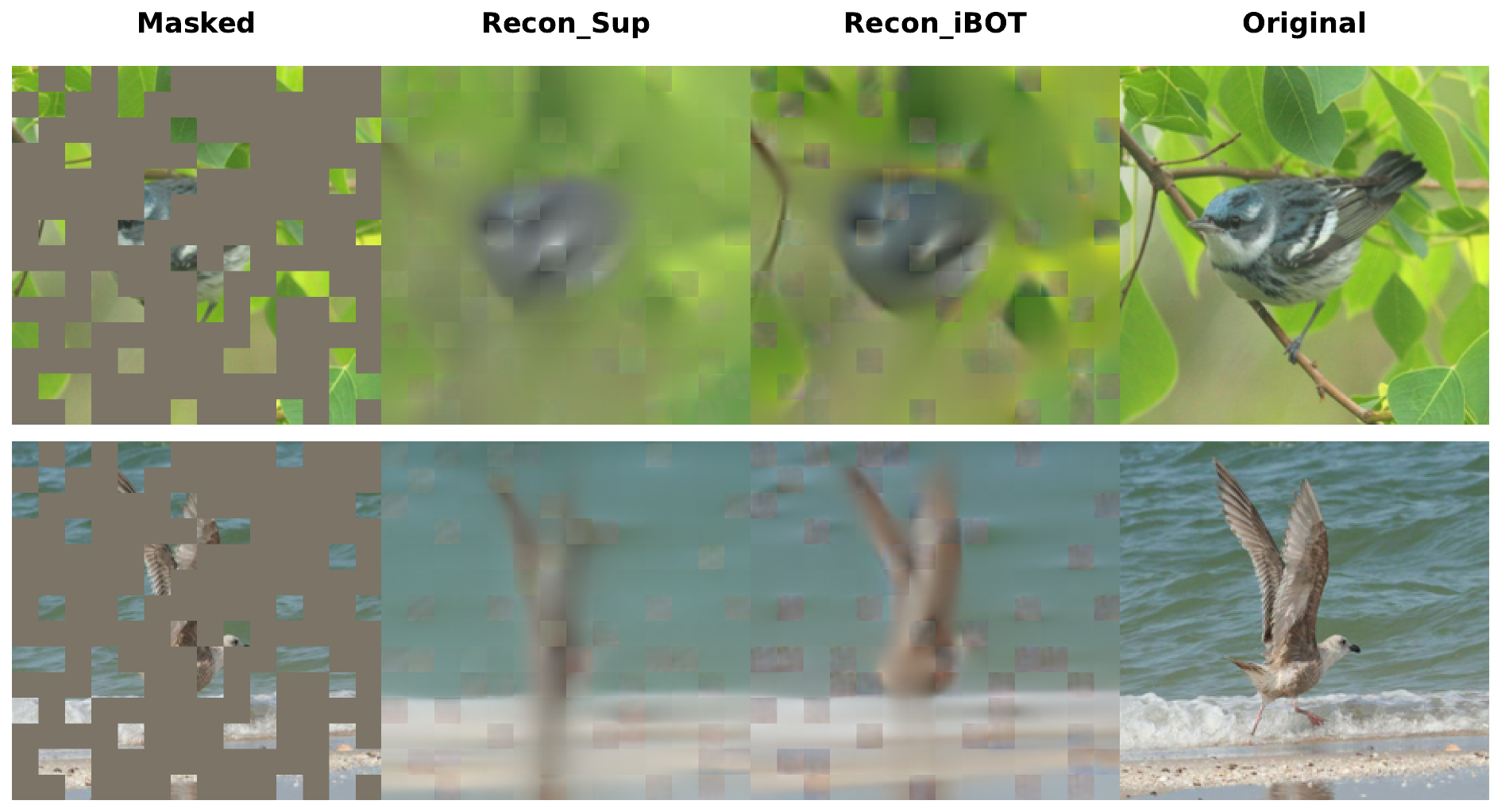}
    \caption{
        \textbf{Uncurated random samples} of CUB-200 images, using an MAE trained on ImageNet-100 with frozen pre-trained encoder.
        For each quadruplet, we show the masked image, Sup-based MAE reconstruction, iBOT-based MAE reconstruction, and the ground-truth.
        The masking ratio is 75\%.
    }
    \label{fig:mae_cub-200}
\end{figure}

\begin{figure}[htp!]
    \centering
    \includegraphics[width=0.95\linewidth]{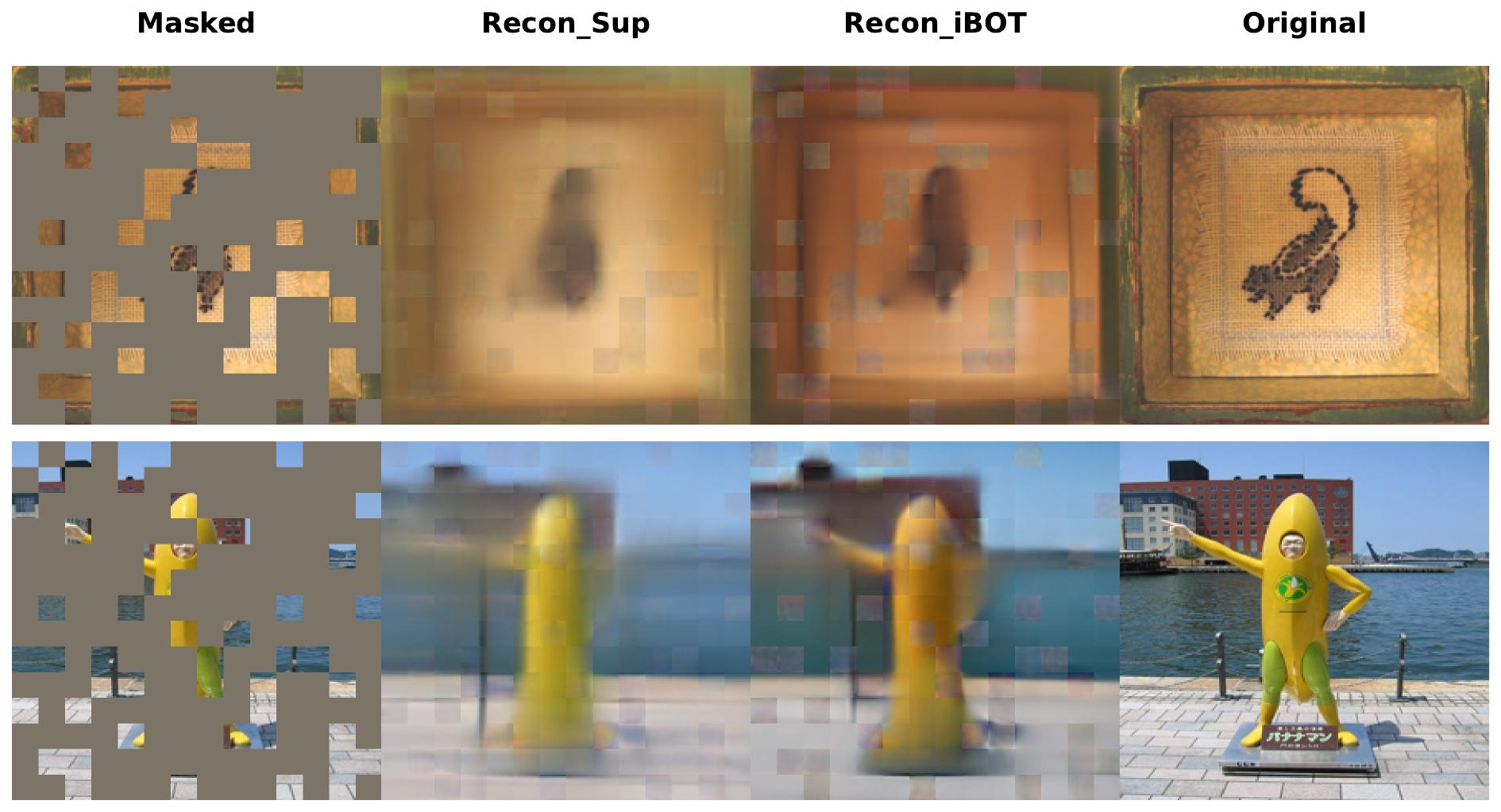}
    \caption{
        \textbf{Uncurated random samples} of ImageNet-R images, using an MAE trained on ImageNet-100 with frozen pre-trained encoder.
        For each quadruplet, we show the masked image, Sup-based MAE reconstruction, iBOT-based MAE reconstruction, and the ground-truth.
        The masking ratio is 75\%.
    }
    \label{fig:mae_imagenet-r}
\end{figure}

Specifically, the ImageNet-100 dataset~\cite{wu2019bic}, a subset of 100 classes from ImageNet~\cite{deng2009imagenet}, is utilized for pre-training.
We adopt the original MAE pre-training paradigm, which inputs images with 75\% masking into the encoder and performs reconstruction with the decoder.
Our aim is to develop a decoder capable of reconstruction while maintaining the encoder's performance, ensuring fairness in comparisons with other methods.
Therefore, we freeze the encoder during pre-training and optimize only the decoder. 
In the original MAE pre-training, the primary objective is to obtain an encoder that produces representations for recognition.
However, since a pre-trained encoder is already employed, 200 epochs of iterations have already achieved convergence in our experiments.

Notably, in the original paper~\cite{he2022mae}, the reconstruction target is the normalized pixels of each masked patch, which improves representation but reduces reconstruction quality.
To focus on reconstruction, we use unnormalized pixels as the target.

Figure~\ref{fig:mae_cub-200} and Figure~\ref{fig:mae_imagenet-r} show the reconstruction results of the Sup-based MAE and the iBOT-based MAE on CUB-200 and ImageNet-R datasets. 
From the reconstruction perspective, the iBOT-based MAE, which uses a self-supervised pre-trained encoder, outperforms the Sup-based MAE.

\begin{table*}[htp!]
    \caption{
        \textbf{Results on 20-task CUB-200}. 
        $\bar{A}$ gives the accuracy averaged over tasks,
        $A_{T}$ gives the final accuracy of all tasks.\\
        \centering We report results over 3 trials.
    }
    \label{tab:cub-200}
    \centering

    \scalebox{0.99}{
        \begin{tabular}{c|c|cc|cc|cc} 
        \hline 
        \multirow{2}{*}{PTM}
        & \multirow{2}{*}{Method}
        & \multicolumn{2}{c|}{$\beta=0.5$} & \multicolumn{2}{c|}{$\beta=0.1$} & \multicolumn{2}{c}{$\beta=0.05$}\\
        \cline{3-8}
        && $\bar{A}$ ($\uparrow$) & $A_{T}$ ($\uparrow)$
        & $\bar{A}$ ($\uparrow$) & $A_{T}$ ($\uparrow)$
        & $\bar{A}$ ($\uparrow$) & $A_{T}$ ($\uparrow)$\\
        \hline
        \hline
        \multirow{4}{*}{Sup}
        &FedL2P     
        & $58.14$\tiny{$\pm 5.19$} & $41.55$\tiny{$\pm 1.07$}
        & $40.20$\tiny{$\pm 3.09$} & $25.31$\tiny{$\pm 0.83$}
        & $33.38$\tiny{$\pm 1.14$} & $22.89$\tiny{$\pm 1.76$}\\
        &FedDualP       
        & $66.16$\tiny{$\pm 2.22$} & $49.63$\tiny{$\pm 0.65$}
        & $46.24$\tiny{$\pm 1.36$} & $30.36$\tiny{$\pm 2.35$}
        & $41.03$\tiny{$\pm 2.65$} & $26.22$\tiny{$\pm 0.83$}\\
        &FedCODA-P        
        & $65.58$\tiny{$\pm 1.93$} & $47.86$\tiny{$\pm 1.26$}
        & $47.97$\tiny{$\pm 2.06$} & $32.08$\tiny{$\pm 1.84$}
        & $43.98$\tiny{$\pm 2.89$} & $27.04$\tiny{$\pm 2.20$}\\
        &\textbf{pMAE}   
        & $\bm{77.18}$\tiny{$\bm{\pm 1.22}$} & $\bm{63.80}$\tiny{$\bm{\pm 0.07}$}
        & $\bm{68.93}$\tiny{$\bm{\pm 1.77}$} & $\bm{55.84}$\tiny{$\bm{\pm 0.84}$}
        & $\bm{66.12}$\tiny{$\bm{\pm 0.75}$} & $\bm{51.85}$\tiny{$\bm{\pm 1.59}$}\\
        \hline
        \hline
        \multirow{4}{*}{iBOT}
        &FedL2P     
        & $14.08$\tiny{$\pm 1.83$} & $6.34$\tiny{$\pm 0.75$}
        & $6.89$\tiny{$\pm 0.47$} & $2.46$\tiny{$\pm 0.7$}
        & $5.53$\tiny{$\pm 0.14$} & $2.5$\tiny{$\pm 0.52$}\\
        &FedDualP       
        & $31.57$\tiny{$\pm 3.83$} & $16.76$\tiny{$\pm 0.95$}
        & $18.06$\tiny{$\pm 1.05$} & $8.46$\tiny{$\pm 1.47$}
        & $15.13$\tiny{$\pm 1.32$} & $7.83$\tiny{$\pm 2.17$}\\
        &FedCODA-P        
        & $38.88$\tiny{$\pm 3.43$} & $23.00$\tiny{$\pm 1.02$}
        & $21.7$\tiny{$\pm 1.12$} & $10.32$\tiny{$\pm 1.83$}
        & $18.7$\tiny{$\pm 1.02$} & $9.86$\tiny{$\pm 1.16$}\\
        &\textbf{pMAE}   
        & $\bm{53.19}$\tiny{$\bm{\pm 1.21}$} & $\bm{39.35}$\tiny{$\bm{\pm 0.78}$}
        & $\bm{43.32}$\tiny{$\bm{\pm 1.93}$} & $\bm{32.55}$\tiny{$\bm{\pm 1.02}$}
        & $\bm{40.69}$\tiny{$\bm{\pm 1.45}$} & $\bm{30.15}$\tiny{$\bm{\pm 0.87}$}\\
        \hline
        \end{tabular}
    }
\end{table*}

\begin{figure*}[htp!]
    \centering
    \includegraphics[width=0.99\textwidth]{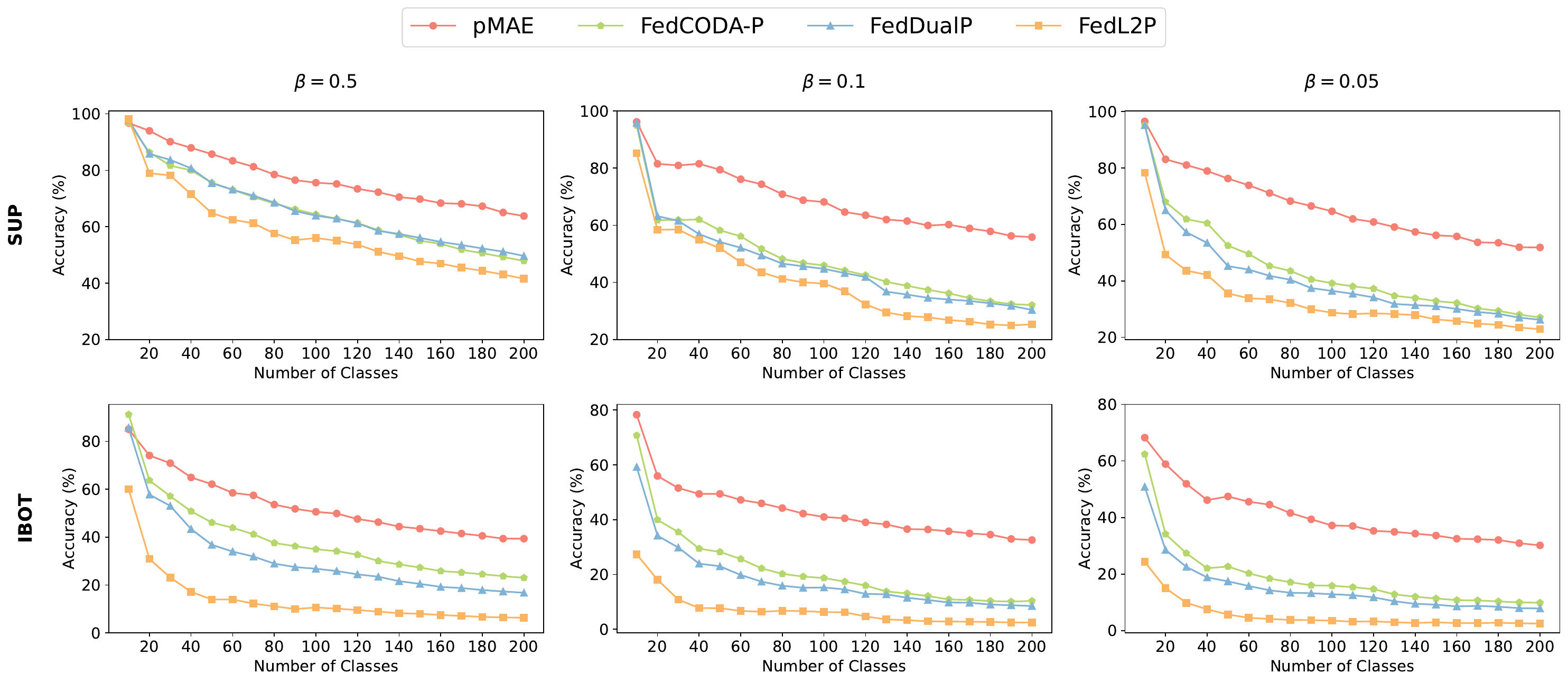}
    \caption{
        Accuracy curves on 20-task CUB-200.
    }
    \label{fig:incacc_cub-200}
\end{figure*}

\subsection{Performance Comparison}

Experiments were conducted on CUB-200 and ImageNet-R datasets with a fixed number of tasks $T=20$, where each task comprises 10 classes.
With a number of clients $K=10$, we varied non-IID degrees $\beta \in \{ 0.5, 0.1, 0.05 \}$ to examine the impact on performance and the non-IID robustness of all methods.
Table~\ref{tab:cub-200} and Table~\ref{tab:imagenet-r} respectively present the results for CUB-200 and ImageNet-R across the average accuracy $\bar{A}$ and last stage accuracy $A_{T}$ metrics.
Figure~\ref{fig:incacc_cub-200} and Figure~\ref{fig:mae_imagenet-r} display the accuracy $A_t$ at each task stage $t$ under different pre-trained models (PTMs) and varying non-IID degrees.

In Table~\ref{tab:cub-200}, our pMAE approach achieves optimal performance levels on the CUB-200 dataset across different non-IID degrees and pre-trained models.
Furthermore, Figure~\ref{fig:incacc_cub-200} illustrates more intuitively that as the degree of non-IID increases, the performance disparity between pMAE and the other methods becomes even more pronounced.

It is noteworthy that when the pre-trained model is based on iBOT, the performance gap between pMAE and the other methods becomes larger compared to the Sup-based model.
This may be attributed to their key-query matching mechanism, which limits performance when using a self-supervised pre-trained model as the backbone.

L2P~\cite{wang2022l2p} introduced a key-query matching mechanism for selecting prompts from a prompt pool to instruct pre-trained transformers.
Building on L2P, DualPrompt~\cite{wang2022dualprompt} enhanced performance by incorporating both general prompts and expert prompts.
CODA-Prompt~\cite{smith2023coda-prompt} utilizes an attention mechanism to construct input-conditioned prompts.

Their commonality lies in relying on a pre-trained encoder to calculate query and construct prompts for classification based on the calculated query,
with the constructed prompts serving a role similar to the discriminative prompt proposed in our pMAE.
Since the construction of prompts is based on calculated query,
these methods heavily depend on the discriminative performance of the pre-trained encoder,
which serves as a frozen feature extractor.

In our experiments, the Sup-based model aligns with the backbone used in~\cite{wang2022l2p,wang2022dualprompt,smith2023coda-prompt}, which exhibits strong discriminative performance.
In contrast, the iBOT-based model, which employs a self-supervised pre-training paradigm, has weaker discriminative ability, resulting in performance degradation.
However, our pMAE directly defines the discriminative prompt without construction, leading to only minimal performance degradation when using the self-supervised pre-trained iBOT-based model.

\begin{table*}[htp!]
    \caption{
        \textbf{Results on 20-task ImageNet-R}. 
        $\bar{A}$ gives the accuracy averaged over tasks,
        $A_{T}$ gives the final accuracy of all tasks.\\
        \centering We report results over 3 trials.
    }
    \label{tab:imagenet-r}
    \centering

    \scalebox{0.99}{
        \begin{tabular}{c|c|cc|cc|cc} 
        \hline 
        \multirow{2}{*}{PTM}
        & \multirow{2}{*}{Method}
        & \multicolumn{2}{c|}{$\beta=0.5$} & \multicolumn{2}{c|}{$\beta=0.1$} & \multicolumn{2}{c}{$\beta=0.05$}\\
        \cline{3-8}
        && $\bar{A}$ ($\uparrow$) & $A_{T}$ ($\uparrow)$
        & $\bar{A}$ ($\uparrow$) & $A_{T}$ ($\uparrow)$
        & $\bar{A}$ ($\uparrow$) & $A_{T}$ ($\uparrow)$\\
        \hline
        \hline
        \multirow{4}{*}{Sup}
        &FedL2P     
        & $48.83$\tiny{$\pm 4.77$} & $41.37$\tiny{$\pm 1.07$}
        & $28.47$\tiny{$\pm 3.53$} & $18.39$\tiny{$\pm 3.64$}
        & $24.99$\tiny{$\pm 2.56$} & $16.01$\tiny{$\pm 3.31$}\\
        &FedDualP       
        & $56.17$\tiny{$\pm 1.84$} & $48.58$\tiny{$\pm 0.65$}
        & $37.25$\tiny{$\pm 0.65$} & $28.61$\tiny{$\pm 2.84$}
        & $31.71$\tiny{$\pm 5.31$} & $21.80$\tiny{$\pm 3.13$}\\
        &FedCODA-P        
        & $59.56$\tiny{$\pm 2.76$} & $\bm{52.40}$\tiny{$\bm{\pm 1.61}$}
        & $39.27$\tiny{$\pm 1.55$} & $30.01$\tiny{$\pm 3.37$}
        & $33.86$\tiny{$\pm 4.63$} & $23.61$\tiny{$\pm 2.13$}\\
        &\textbf{pMAE}   
        & $\bm{60.04}$\tiny{$\bm{\pm 0.17}$} & $50.89$\tiny{$\pm 1.36$}
        & $\bm{53.55}$\tiny{$\bm{\pm 2.10}$} & $\bm{46.31}$\tiny{$\bm{\pm 2.26}$}
        & $\bm{46.72}$\tiny{$\bm{\pm 3.45}$} & $\bm{39.98}$\tiny{$\bm{\pm 1.75}$}\\
        \hline
        \hline
        \multirow{4}{*}{iBOT}
        &FedL2P     
        & $21.18$\tiny{$\pm 2.06$} & $22.14$\tiny{$\pm 0.53$}
        & $10.45$\tiny{$\pm 0.83$} & $8.25$\tiny{$\pm 1.01$}
        & $9.92$\tiny{$\pm 1.08$} & $6.86$\tiny{$\pm 0.11$}\\
        &FedDualP       
        & $41.90$\tiny{$\pm 1.94$} & $35.26$\tiny{$\pm 1.55$}
        & $24.71$\tiny{$\pm 0.51$} & $18.12$\tiny{$\pm 2.30$}
        & $21.90$\tiny{$\pm 2.06$} & $14.66$\tiny{$\pm 1.47$}\\
        &FedCODA-P        
        & $50.28$\tiny{$\pm 2.46$} & $43.08$\tiny{$\pm 0.77$}
        & $30.34$\tiny{$\pm 1.00$} & $22.44$\tiny{$\pm 2.41$}
        & $26.53$\tiny{$\pm 2.81$} & $17.53$\tiny{$\pm 2.16$}\\
        &\textbf{pMAE}   
        & $\bm{53.47}$\tiny{$\bm{\pm 0.79}$} & $\bm{46.90}$\tiny{$\bm{\pm 0.51}$}
        & $\bm{47.78}$\tiny{$\bm{\pm 0.72}$} & $\bm{43.13}$\tiny{$\bm{\pm 0.52}$}
        & $\bm{42.65}$\tiny{$\bm{\pm 3.44}$} & $\bm{38.86}$\tiny{$\bm{\pm 0.59}$}\\
        \hline
        \end{tabular}
    }
\end{table*}

\begin{figure*}[htp!]
    \centering
    \includegraphics[width=0.99\textwidth]{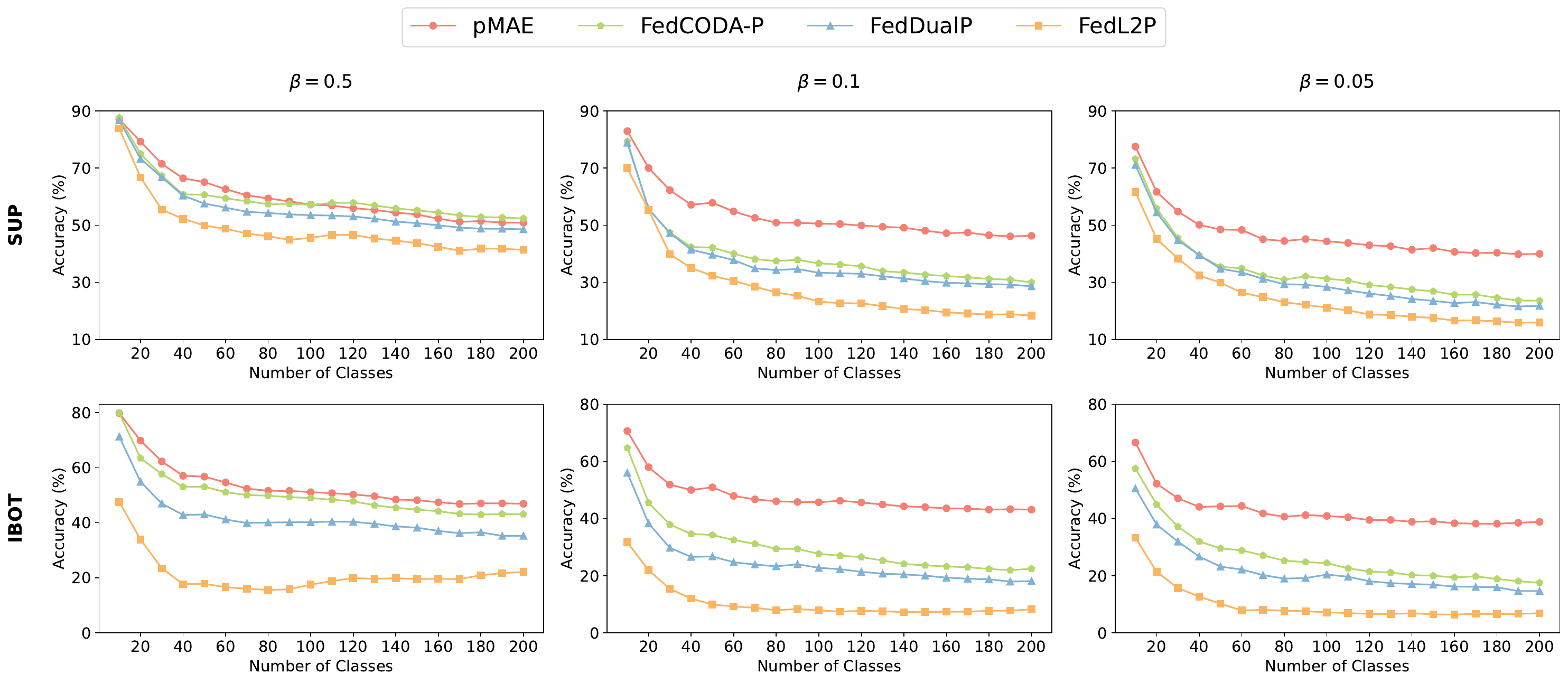}
    \caption{
        Accuracy curves on 20-task ImageNet-R.
    }
    \label{fig:incacc_imagenet-r}
\end{figure*}

For ImageNet-R dataset, as shown in Table~\ref{tab:imagenet-r} and Figure~\ref{fig:incacc_imagenet-r},
when using the Sup-based model with $\beta=0.5$ (indicating a relatively low non-IID degree),
pMAE achieves an average accuracy $\bar{A}$ comparable to FedCODA-P and FedDualP, while FedCODA-P achieves a higher final accuracy $A_t$.
This may be attributed to the out-of-distribution characteristic of ImageNet-R dataset.
FedCODA-P and FedDualP utilize specific prompts to retain knowledge acquired from each task.
Consequently, they are better suited to handle the out-of-distribution data in ImageNet-R,
allowing for both the learning and preserving of each task's distinct distribution.
In contrast, pMAE employs a single discriminative prompt, which may limit its capability to handle the out-of-distribution data in ImageNet-R.

Although the out-of-distribution data in ImageNet-R limits the effectiveness of the discriminative prompt,
the mechanism of using restore information for reconstruction on the server to fine-tune the discriminative prompt and classifier parameters
enables pMAE to effectively mitigate performance degradation caused by catastrophic forgetting and non-IID issues, allowing it to achieve optimal performance in other settings.
Furthermore, as the non-IID degree increases, the performance gap between pMAE and other methods becomes more significant, consistent with the observations from CUB-200.

\subsection{Enhancing Other Methods with pMAE}

The generality of pMAE enables seamless integration with existing prompt-based methods~\cite{wang2022l2p, wang2022dualprompt, smith2023coda-prompt} to enhance performance.
Since these methods already utilize an encoder for feature extraction, integrating pMAE simply requires adding a decoder to establish a complete MAE.
During client training, the reconstructive prompt is optimized with MSE loss (as outlined in Section~\ref{sec:method}).
Restore information is then extracted on the client, uploaded to the server, and used with the reconstructive prompt for reconstruction.
These reconstructed images are subsequently used for fine-tuning parameters tailored for classification, without any further modifications.

\begin{figure}[htp!]
    \centering
    \includegraphics[width=0.95\linewidth]{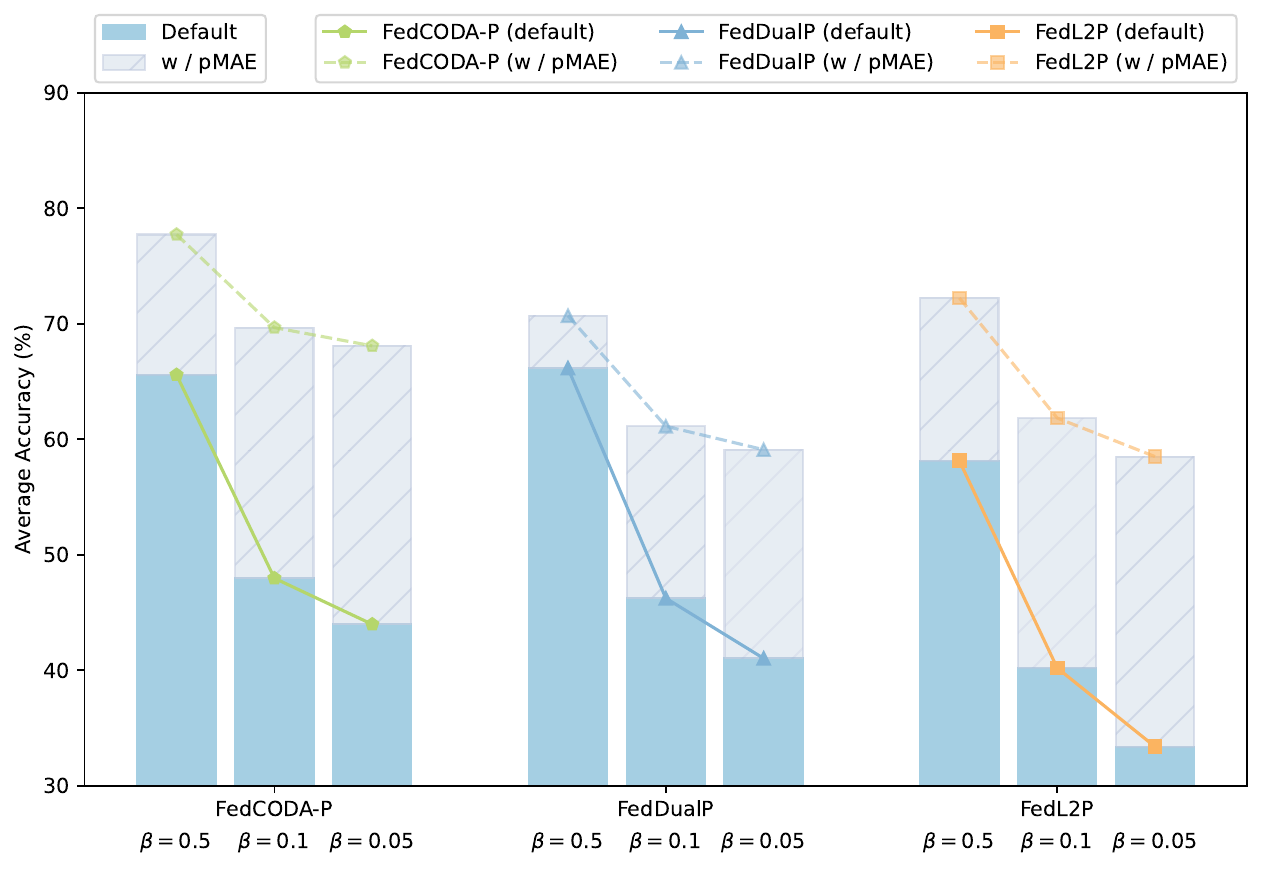}
    \caption{
        \textbf{Sup-based} improvement of other methods with pMAE for average accuracy $\bar{A}$ on 20-task CUB-200.
    }
    \label{fig:w_pmae_cub-200_sup}
\end{figure}

\begin{figure}[htp!]
    \centering
    \includegraphics[width=0.95\linewidth]{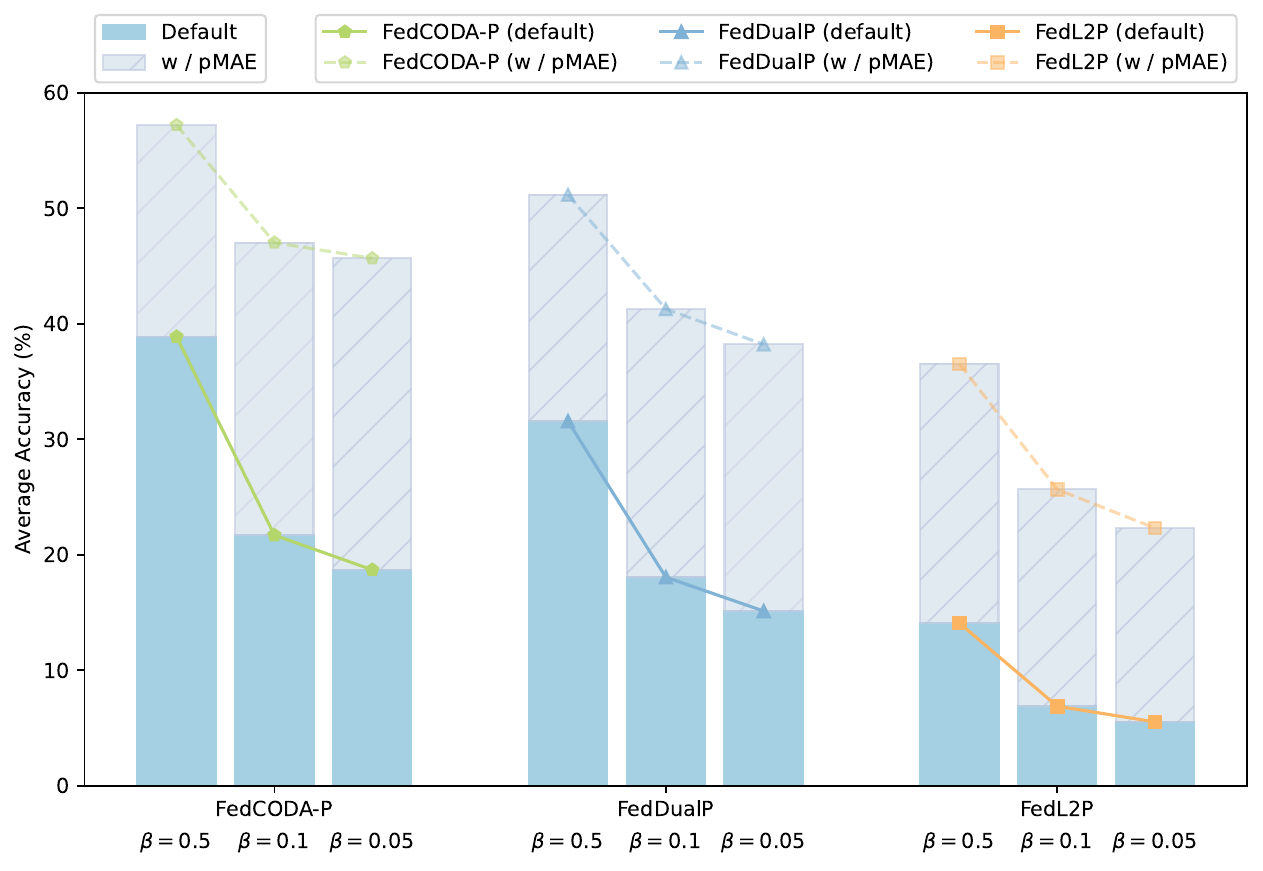}
    \caption{
        \textbf{iBOT-based} improvement of other methods with pMAE for average accuracy $\bar{A}$ on 20-task CUB-200.
    }
    \label{fig:w_pmae_cub-200_ibot}
\end{figure}

The experimental results, presented in Figure~\ref{fig:w_pmae_cub-200_sup} and Figure~\ref{fig:w_pmae_cub-200_ibot},
show that FedCODA-P, FedDualP, and FedL2P achieve substantial performance improvements across different pre-trained models and non-IID degrees,
especially when using an iBOT-based model.
It is noteworthy that the y-axis starting points in Figure~\ref{fig:w_pmae_cub-200_sup} and Figure~\ref{fig:w_pmae_cub-200_ibot} differ.
Despite the bars appear lower in Figure~\ref{fig:w_pmae_cub-200_sup}, the Sup-based model actually performs better than the iBOT-based model.

Additionally, as indicated by the slope, integrating pMAE reduces the rate of performance decline as non-IID degree increases,
underscoring pMAE's contribution to robustness.

\subsection{Ablation Study}

To better understand the effects of individual components of pMAE, we perform an ablation study on CUB-200 dataset.
Table~\ref{tab:ablation-sup} and Table~\ref{tab:ablation-ibot} offer insights into the impact of
ablating reconstructive prompt, reducing restoration information, and removing restoration pool.

\begin{figure}[htp!]
    \centering
    \includegraphics[width=0.95\linewidth]{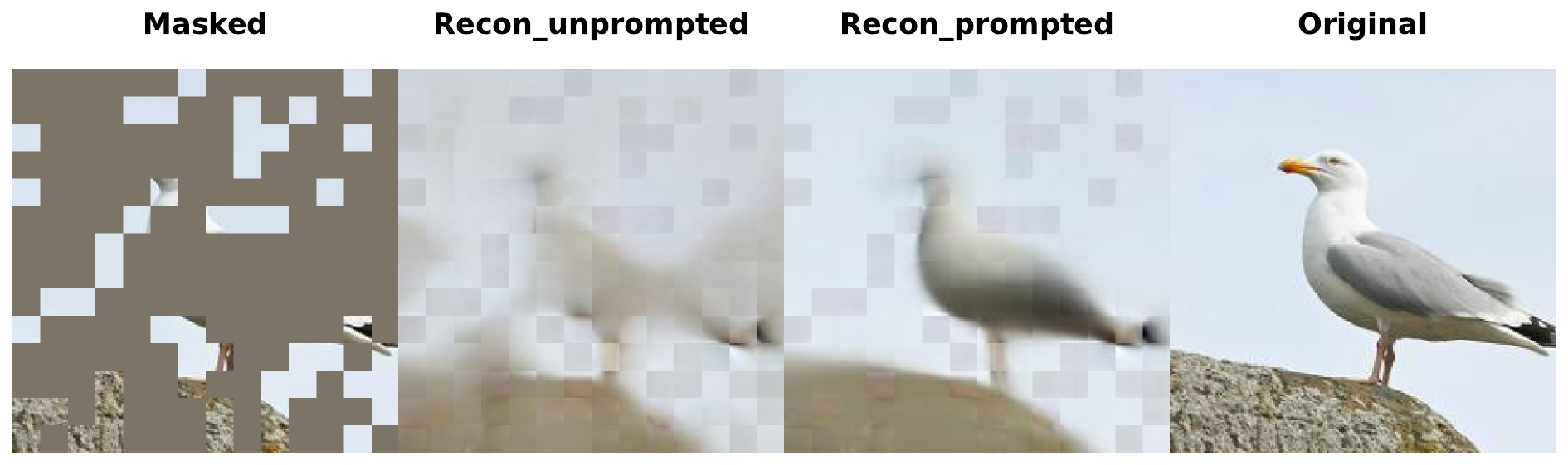}
    \caption{
        \textbf{Sup-based} reconstruction of an example from CUB-200 images.
        For each quadruplet, we show the masked image, unprompted reconstruction, prompted reconstruction, and the ground-truth.
        The masking ratio is 75\%.
    }
    \label{fig:wo_prompt_mae_sup}
\end{figure}

\begin{figure}[htp!]
    \centering
    \includegraphics[width=0.95\linewidth]{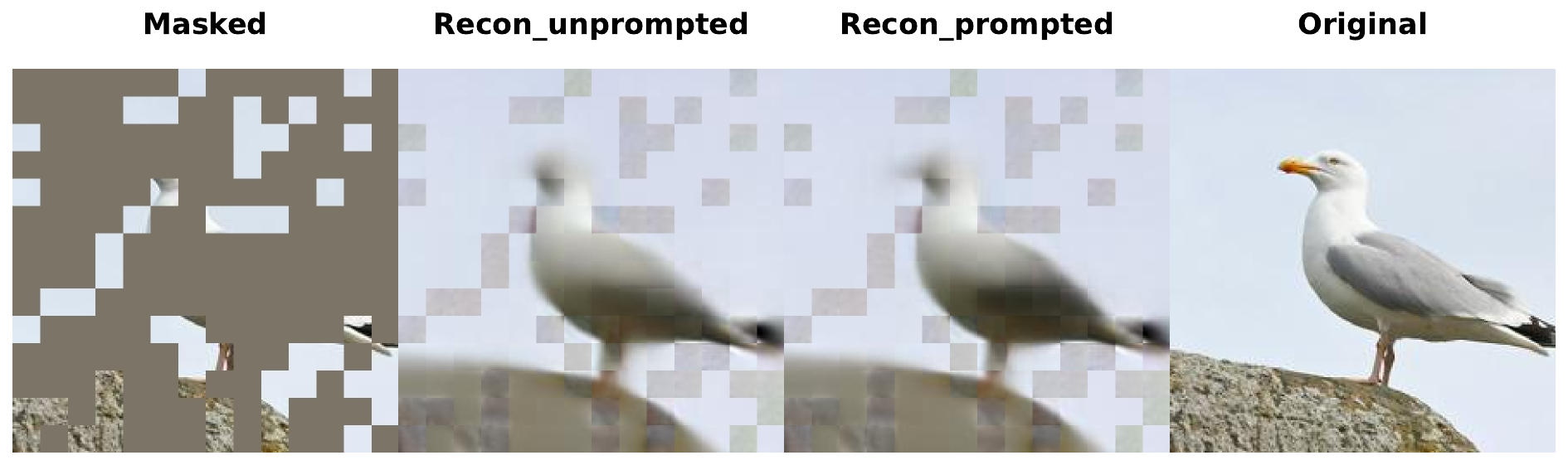}
    \caption{
        \textbf{iBOT-based} reconstruction of an example from CUB-200 images.
        For each quadruplet, we show the masked image, unprompted reconstruction, prompted reconstruction, and the ground-truth.
        The masking ratio is 75\%.
    }
    \label{fig:wo_prompt_mae_ibot}
\end{figure}

Specifically, the ablation of reconstructive prompt results in a decline in the quality of image reconstructed on the server side.
As shown in Figures~\ref{fig:wo_prompt_mae_sup} and~\ref{fig:wo_prompt_mae_ibot},
image reconstructed with reconstructive prompt exhibit superior quality,
which is crucial for the fine-tuning process on the server side.
The reconstructive prompt offers a more significant quality improvement for the Sup-based model compared to the iBOT-based model,
which can also be confirmed by the performance degradation resulting from ablating reconstructive prompt in Tables~\ref{tab:ablation-sup} and~\ref{tab:ablation-ibot}.

The default number of uploaded restore information is set to $u=4$,
we reduce it to $u=1$ to assess its impact on the performance of pMAE.
It is observed that reducing restore information resulted in a more pronounced drop in average accuracy compared to ablating reconstructive prompt.

\begin{table}[htp!]
    \caption{
        \textbf{Sup-based} ablation results for average accuracy $\bar{A}$ on 20-task CUB-200.
    }
    \label{tab:ablation-sup}
    \centering

    \scalebox{0.88}{
        \begin{tabular}{c||c|c|c} 
        \hline 
        Method & $\beta=0.5$ & $\beta=0.1$ & $\beta=0.05$\\
        \hline
        \textbf{pMAE}   
        & $\bm{77.18}$\tiny{$\bm{\pm 1.22}$}
        & $\bm{68.93}$\tiny{$\bm{\pm 1.77}$}
        & $\bm{66.12}$\tiny{$\bm{\pm 0.75}$}\\
        \hline
        Ablate Reconstructive Prompt
        & $74.19$\tiny{$\pm 2.49$}
        & $65.27$\tiny{$\pm 2.28$}
        & $62.21$\tiny{$\pm 4.00$}\\
        Reduce Restore Information
        & $68.74$\tiny{$\pm 2.66$}
        & $57.16$\tiny{$\pm 4.15$}
        & $55.00$\tiny{$\pm 2.70$}\\
        Remove Restore Pool
        & $65.36$\tiny{$\pm 2.36$}
        & $54.45$\tiny{$\pm 3.91$}
        & $51.20$\tiny{$\pm 4.86$}\\
        \hline
        \end{tabular}
    }
\end{table}

\begin{table}[htp!]
    \caption{
        \textbf{iBOT-based} ablation results for average accuracy $\bar{A}$ on 20-task CUB-200.
    }
    \label{tab:ablation-ibot}
    \centering

    \scalebox{0.88}{
        \begin{tabular}{c||c|c|c} 
        \hline 
        Method & $\beta=0.5$ & $\beta=0.1$ & $\beta=0.05$\\
        \hline
        \textbf{pMAE}   
        & $\bm{53.19}$\tiny{$\bm{\pm 1.21}$}
        & $\bm{43.32}$\tiny{$\bm{\pm 1.93}$}
        & $\bm{40.69}$\tiny{$\bm{\pm 1.45}$}\\
        \hline
        Ablate Reconstructive Prompt
        & $52.92$\tiny{$\pm 1.42$}
        & $43.15$\tiny{$\pm 2.09$}
        & $40.39$\tiny{$\pm 1.14$}\\
        Reduce Restore Information
        & $47.25$\tiny{$\pm 2.30$}
        & $35.51$\tiny{$\pm 1.92$}
        & $31.74$\tiny{$\pm 1.05$}\\
        Remove Restore Pool
        & $39.06$\tiny{$\pm 3.90$}
        & $27.44$\tiny{$\pm 1.05$}
        & $27.77$\tiny{$\pm 2.51$}\\
        \hline
        \end{tabular}
    }
\end{table}

Remarkably, the removal of restore pool incurs a significant decline in average accuracy,
emphasizing its pivotal role in retaining distributions from old task data,
which is a major factor leading to catastrophic forgetting in FCL.

\section{Conclusion}
\label{sec:conclusion}

In this work, we propose to use masked autoencoders (MAEs) as parameter-efficient federated continual learners, called \textbf{pMAE}.
Our pMAE effectively addresses the challenges of catastrophic forgetting and non-IID issues by employing the image reconstruction ability of MAEs.
Specifically, on the client side, pMAE learns a reconstructive prompt through reconstruction.
On the server side, it aggregates uploaded restore information and reconstructs images with the reconstructive prompt to capture data distributions across tasks and clients.
Experiments conducted on benchmark datasets demonstrate the effectiveness of pMAE, showing competitive performance compared to existing prompt-based methods and highlighting its ability to enhance their capabilities,
particularly when using self-supervised pre-trained transformers.

\bibliographystyle{plainnat}
% \bibliography{IEEE_Trans}
\bibliography{egbib}

\end{document}